\documentclass[runningheads]{llncs}
\usepackage{soul}
\usepackage{amsmath}
\usepackage{amssymb}
\usepackage{mathtools}
\usepackage{amsfonts}
\usepackage{commath}
\usepackage{bm}
\usepackage{dsfont}
\usepackage{bbm}
\usepackage[misc]{ifsym}
\usepackage{float}

\usepackage{hyperref}
\usepackage{cleveref}
\usepackage{cite}
\usepackage{listings}
\usepackage{svg}
\usepackage{syntonly}
\usepackage{graphicx}
\usepackage{subcaption}
\usepackage[inline]{enumitem}
\usepackage{tabularx}
\usepackage{hhline}

\usepackage[T1]{fontenc}
\usepackage[utf8]{inputenc}
\usepackage[english]{babel}
\usepackage{csquotes}
\MakeOuterQuote{"}
\usepackage{xcolor}

\DeclareTextFontCommand{\textsymb}{\symb}

\usepackage{booktabs}
\usepackage{multirow}

\newcommand{\increment}{\mathbf{\Delta}}
\renewcommand{\bfseries}{\fontseries{b}\selectfont}
\newrobustcmd{\B}{\bfseries}

\newcommand{\etal}{\textit{et al}.}

\begin{document}

\titlerunning{Knowledge distillation with Segment Anything (SAM) model}
\title{Knowledge distillation with Segment Anything (SAM) model for Planetary Geological Mapping}
%
%

\author{Sahib Julka\inst{1}\orcidID{0000-0003-3566-5507} \and
Michael Granitzer\inst{1}\orcidID{0000-0002-2952-5519}
}
%
%
\institute{Chair of Data Science, University of Passau, 94036 Passau, Germany \\
\email{\{sahib.julka, michael.granitzer\}@uni-passau.de}\\}

\maketitle              
\begin{abstract}

Planetary science research involves analysing vast amounts of remote sensing data, which are often costly and time-consuming to annotate and process.  One of the essential tasks in this field is geological mapping, which requires identifying and outlining regions of interest in planetary images, including geological features and landforms. However, manually labelling these images is a complex and challenging task that requires significant domain expertise and effort. To expedite this endeavour, we propose the use of knowledge distillation using the recently introduced cutting-edge Segment Anything (SAM) model. We demonstrate the effectiveness of this prompt-based foundation model for rapid annotation and quick adaptability to a prime use case of mapping planetary skylights. Our work reveals that with a small set of annotations obtained with the right prompts from the model and subsequently training a specialised domain decoder, we can achieve satisfactory semantic segmentation on this task. Key results indicate that the use of knowledge distillation can significantly reduce the effort required by domain experts for manual annotation and improve the efficiency of image segmentation tasks. This approach has the potential to accelerate extra-terrestrial discovery by automatically detecting and segmenting Martian landforms. 

\end{abstract}
\section{Introduction}
\label{sec: Intro}
We have recently witnessed a paradigm shift in AI with the advent of \emph{foundation models} utilising astronomical amounts of data. The fields of natural language processing and multi-modal learning have been revolutionised with the emergence of ChatGPT and the like~\cite{lund2023chatting, qin2023chatgpt}. The very first foundation models such as CLIP~\cite{radford2021learning}, ALIGN~\cite{jia2021scaling}, and DALLE~\cite{ramesh2021zero}, have focused on pre-training approaches but are not suited to image segmentation. However, recently, Segment Anything (SAM)~\cite{kirillov2023segment} was released, which is a large vision transformer ViT-based~\cite{dosovitskiy2020image} model trained on the large visual corpus (SA-1B) containing more than 11 million images and one billion masks. SAM is designed to generate a valid segmentation result for any prompt. However, SAM is trained on general world case scenarios with popular structures. 
Recent studies have revealed that SAM can fail on typical medical image segmentation tasks~\cite{deng2023segment,hu2023sam} and other challenging scenarios~\cite{chen2023learning,tang2023can,ji2023segment,ji2023sam}. Since SAM’s training set mainly contains natural image datasets, it may not be directly transferable to niche tasks on data such as magnetic resonance (MRI), or HiRISE imaging~\footnote{``High-Resolution Imaging Science Experiment'' is camera aboard the Mars Reconnaissance Orbiter (MRO) spacecraft, which is designed to capture high-resolution images of the Martian surface and provide detailed information about the planet's geology and atmosphere.}, amongst other specialised data formats. Nonetheless, SAM is still a powerful tool that has a powerful image encoder and its prompt functionality can significantly boost the efficiency of manual annotation. In the planetary science domain, where vast amounts of remote sensing data are gathered, annotation is an intensive task. An approach that reduces the effort on the domain experts' end is highly desired. In these scenarios, active learning~\cite{julka2023deep, julka2022active} and knowledge distillation~\cite{gou2021knowledge} via training a specialised model with relatively fewer samples can be highly valuable.

\subsection{Segment Anything Model}

SAM utilises a vision transformer-based~\cite{jaderberg2015spatial} approach to extract image features and prompt encoders to incorporate user interactions for segmentation tasks. The extracted image features and prompt embeddings are then processed by a mask decoder to generate segmentation results and confidence scores. There are four~\footnote{Text prompt is currently not released.} types of prompts supported by SAM, namely \emph{point}, \emph{text}, \emph{box}, and \emph{mask} prompts.

For the points prompt, SAM encodes each point with Fourier positional encoding and two learnable tokens that specify foreground and background. The bounding box prompt is encoded by using the point encoding of its top-left and bottom-right corners. SAM employs a pre-trained text encoder in CLIP for encoding the free-form text prompt. The mask prompt has the same spatial resolution as the input image and is encoded by convolution feature maps.

Finally, SAM's mask decoder consists of two transformer layers with a dynamic mask prediction head and an Intersection-over-Union (IoU) score regression head. The mask prediction head generates three downscaled masks, corresponding to the whole object, part, and subpart of the object. SAM supports three main segmentation modes: fully automatic, bounding box, and point mode.

\begin{figure}[h!tb]
    \centering
        \includegraphics[width =\linewidth]{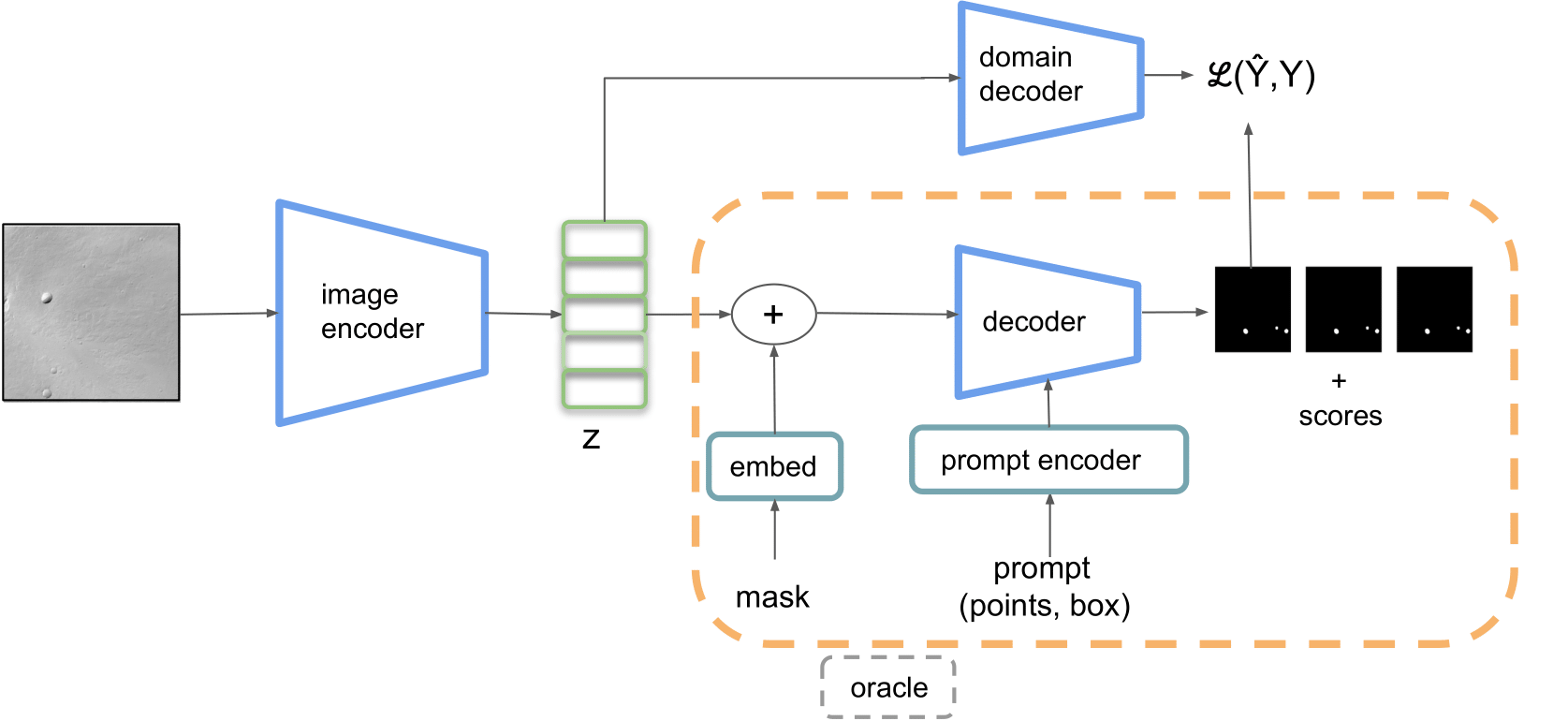}
        \caption{Overview of our deployed approach by extending SAM. It consists of SAMs image encoder that learns an embedding of the image, and a specialised decoding unit to learn the domain-specific semantics. SAMs prompt encoder and mask decoder, represented within the orange bounding box are utilised only for annotating incrementally the $\increment (\mathbb{N})$ training samples. While training the domain decoder, the image encoder is frozen so as not to update its weights.}
    \label{fig: architecture}
\end{figure}

\subsection{Landform detection on Mars using HiRISE images}
\begin{figure}[h!tb]
  \centering
  \begin{subfigure}[b]{0.15\textwidth}
    \includegraphics[width=\textwidth]{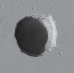}
    \caption{Type 1a}
  \end{subfigure}
  \begin{subfigure}[b]{0.15\textwidth}
    \includegraphics[width=\textwidth]{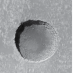}
    \caption{Type 1b}
  \end{subfigure}
  \begin{subfigure}[b]{0.15\textwidth}
    \includegraphics[width=\textwidth]{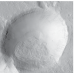}
    \caption{Type 2a}
  \end{subfigure}
  \begin{subfigure}[b]{0.15\textwidth}
    \includegraphics[width=\textwidth]{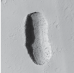}
    \caption{Type 2b}
  \end{subfigure}
  \begin{subfigure}[b]{0.15\textwidth}
    \includegraphics[width=\textwidth]{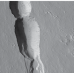}
    \caption{Type 3}
  \end{subfigure}
  \begin{subfigure}[b]{0.15\textwidth}
    \includegraphics[width=\textwidth]{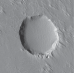}
    \caption{Type 4}
  \end{subfigure}
  \caption{Principal types of pits and skylights found on Mars terrain: (a) Skylight with possible cave entrance (Type 1a). (b) Pit with possible relation to cave entrance (Type 1b). (c) ``Bowl'' pit with a possible connection to lava tubes (Type 2a). (d) Pit with uncertain connection to lava tubes or dikes (Type 2b). (e) Coalescent pits (Type 3). (f) Pit with a possible connection to lava tubes (Type 4)~\cite{nodjoumi2023deeplandforms}. }
  \label{fig: landform_types}
\end{figure}

Mapping planetary landforms plays a crucial role in various tasks such as surveying, environmental monitoring, resource management, and planning. On Earth, for example, the presence of water triggers several geological and geomorphological processes~\cite{allemand2011thirty}. Conversely, on Mars, researchers have found correlations between the presence of certain landforms such as pits, sinkholes, and landslides and the possible presence of water~\cite{baker2001water, baker2006geomorphological}. However, identifying, classifying, and drawing regions of interest manually is a complex and time-consuming process~\cite{nodjoumi2023deeplandforms}-- one that would greatly benefit from automation.

In this regard, the identification and segmentation of various Martian landforms have gained increasing attention in recent years~\cite{nodjoumi2023deeplandforms,julka2021generative, jiang2022automated, palafox2017automated}.~\Cref{fig: landform_types} shows an overview of some of the pits and skylights that can be identified on the Martian terrain. In this study, we focus only on these landforms, utilising a dataset prepared exclusively for it (cf.~\Cref{sec: method_dataset}). Automatic detection and segmentation of these landforms have the potential to accelerate the identification of potential landing sites for future missions, study the geological history of Mars, and contribute to a better understanding of the planet's potential habitability. Therefore, this endeavour is of significant importance in planetary science.

\section{Method}
\label{sec: method}
\subsection{Dataset}
\label{sec: method_dataset}
The data used in this work are images acquired by image sensors operating in the visible (VIS) and Near InfraRed (NIR) spectrums on board probes orbiting Mars. This data set is composed of images by HiRISE instrument and downloaded both as Reduced Data Record (RDR)
and Experiment Data Record (EDR) format from public space archives such as PDS Geosciences Node Orbital Data Explorer (ODE)~\footnote{https://ode.rsl.wustl.edu/}. With this, Nodjoumi \etal~\cite{nodjoumi2023deeplandforms} released a processed dataset with 486 samples. This dataset is split into 405 images for training, 25 for validation and the rest for testing. 
In their work, they train a Mask-RCNN using all images annotated manually.
In order to explore the applicability of knowledge distillation, we incrementally select train samples for annotation and subsequently train the domain decoder with these. This, in effect, is analogous to learning correct prompts for the task, with the least amount of annotated samples. 

\begin{figure}[h!tb]
  \centering
  \begin{subfigure}[b]{0.23\textwidth}
    \includegraphics[width=\textwidth]{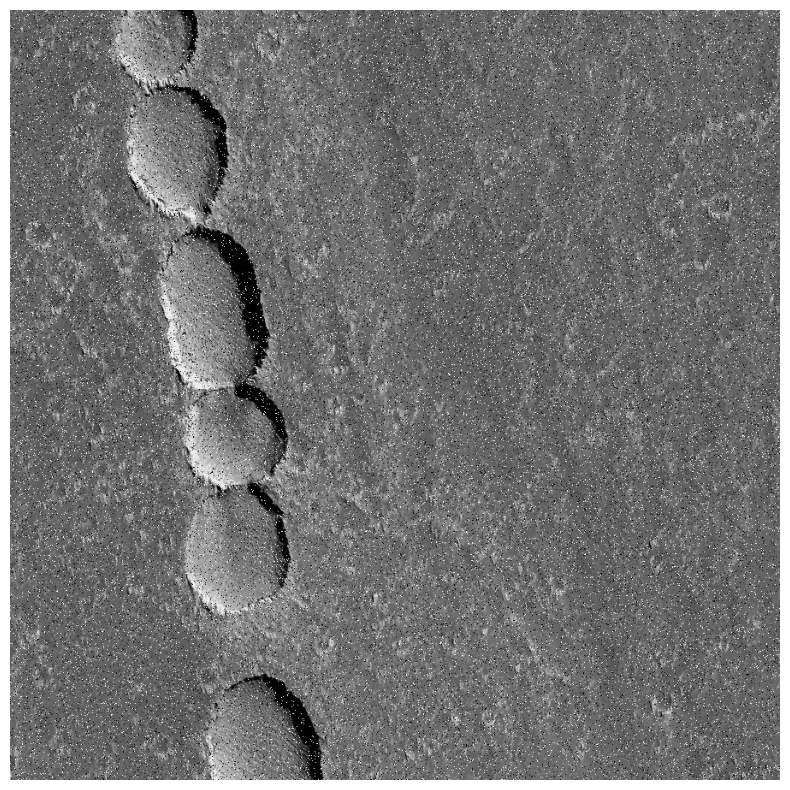}
    \caption{Image}
  \end{subfigure}
  \begin{subfigure}[b]{0.23\textwidth}
    \includegraphics[width=\textwidth]{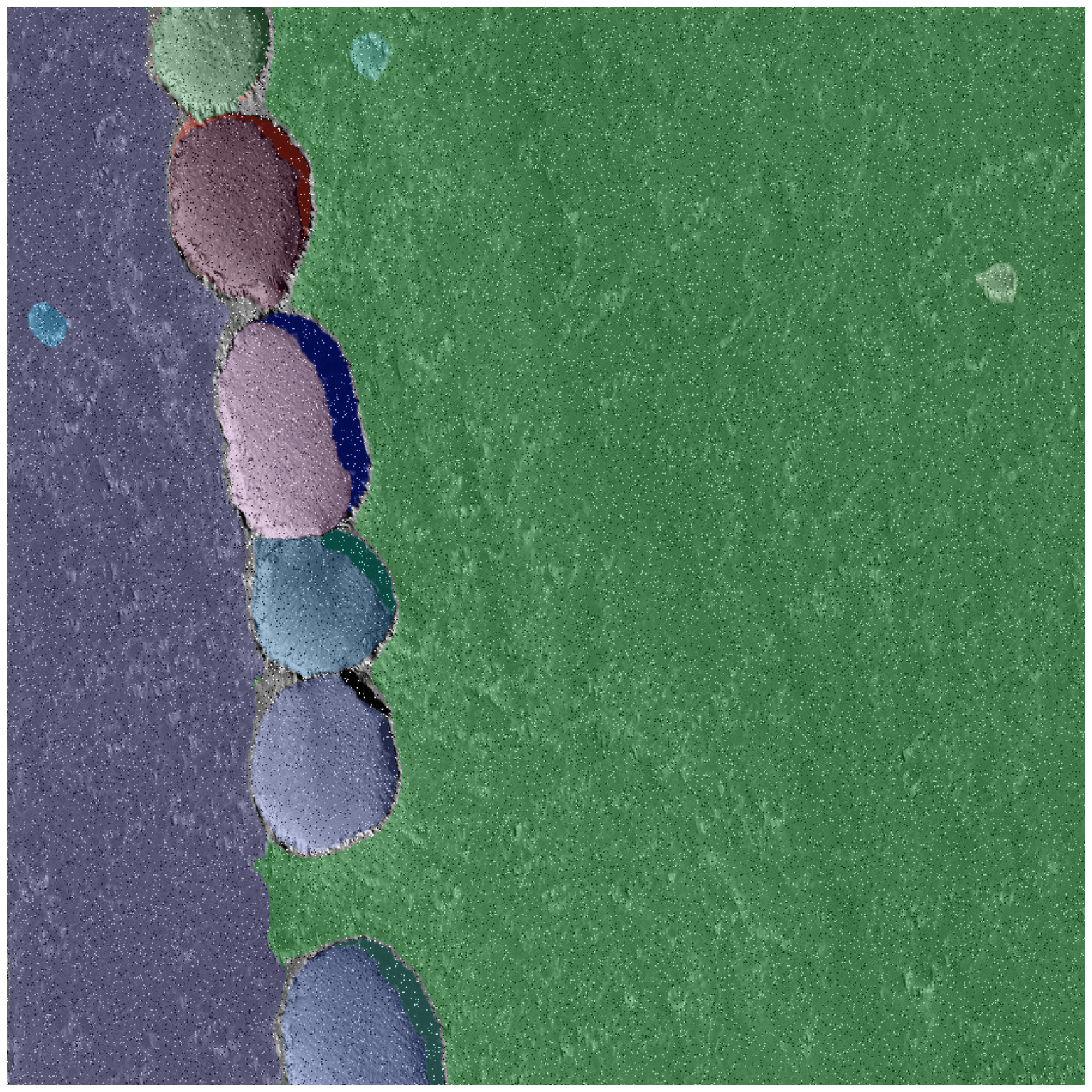}
    \caption{Automatic}
  \end{subfigure}
  \begin{subfigure}[b]{0.23\textwidth}
    \includegraphics[width=\textwidth]{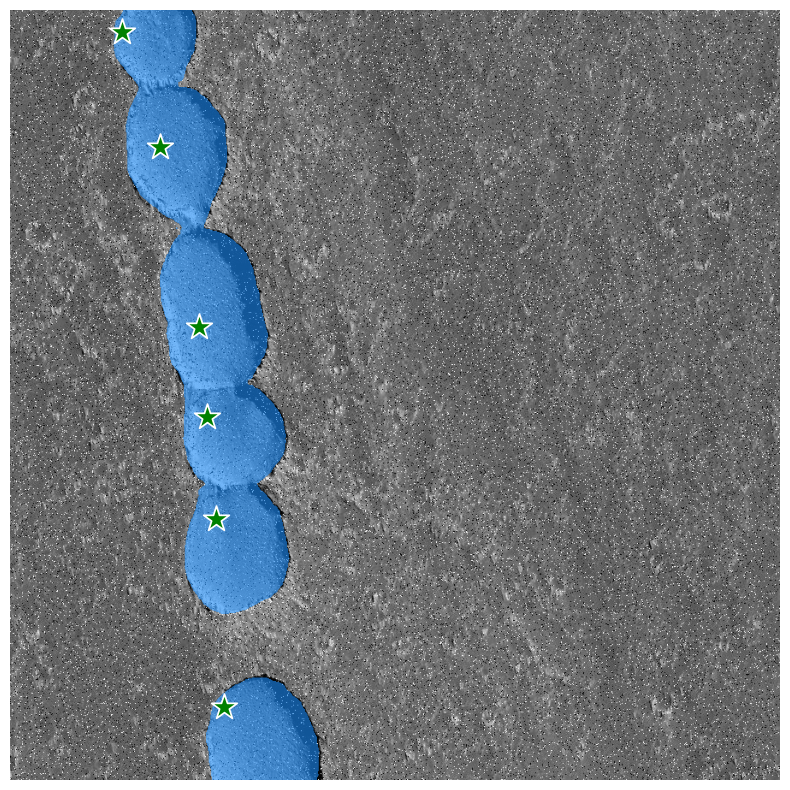}
    \caption{Point prompt}
  \end{subfigure}
  \begin{subfigure}[b]{0.23\textwidth}
    \includegraphics[width=\textwidth]{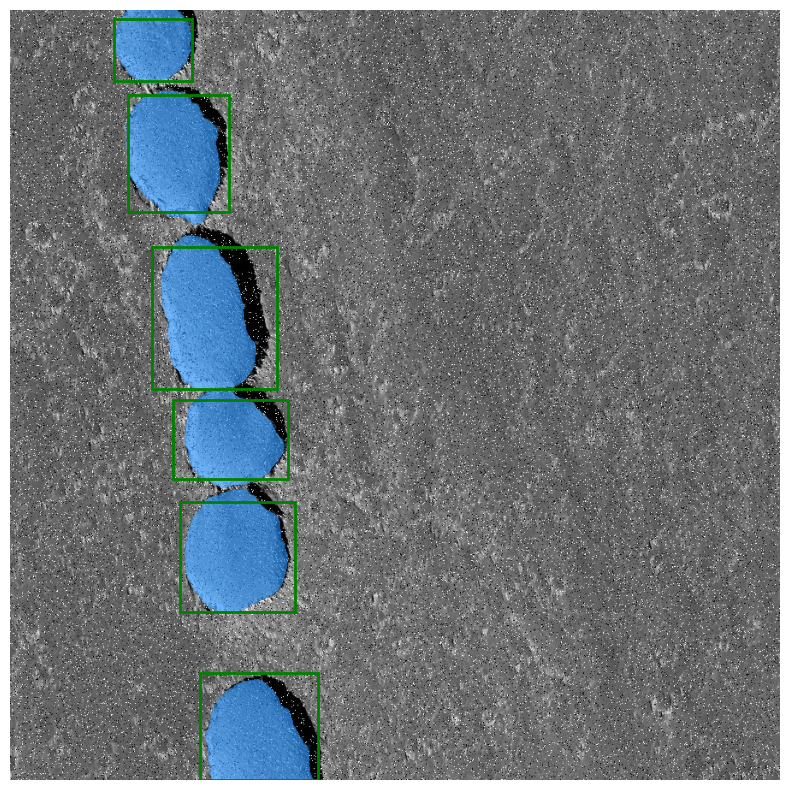}
    \caption{Box prompt}
  \end{subfigure}
  \caption{An overview of generation of segmentation masks with the three different prompt settings in SAM. The box prompt delineates the land mass from the adjacent shadow in comparison to the point prompt. }
  \label{fig: annotation}
\end{figure}

\subsection{Prompt-mode selection for annotation}

We conducted an evaluation of the SAM model using the three different prompt settings:
\begin{enumerate*}[label=(\alph*)]
\item In the automatic prompt setting, SAM generates single-point input prompts in a grid pattern across the input image and selects high-quality masks using non-maximal suppression. All parameters were set to their default values. In the case of multiple masks being obtained, we selected the mask with the highest returned IoU score.
\item In the point prompt setting, we used the centre of the ground truth regions of interest as the point prompts for SAM.
\item  In the box prompt setting, we computed the bounding box for SAM around the ground-truth mask.\end{enumerate*}~\Cref{fig: annotation} illustrates the mask generation on an exemplary sample for the three modes. Clearly, the automatic prompt simply segments all regions in a semantic agnostic way. Point and box prompts generate high-quality segmentation masks, with an average image level IoU above 90~\%. Although in our case, point and box prompt performed relatively comparably on simpler cases, we empirically found box prompt to be most reliable in occluded and shadowy scenes and thus chose that to be used for final annotations. In practice, the expert would need only a few seconds to draw boxes around all relevant regions of interest on a sample.

\subsection{Domain Decoder}
\subsubsection{Why not directly fine-tune the SAM decoder?}
\begin{figure}[h!tb]
    \begin{centering}
        \includegraphics[width =0.7\linewidth]{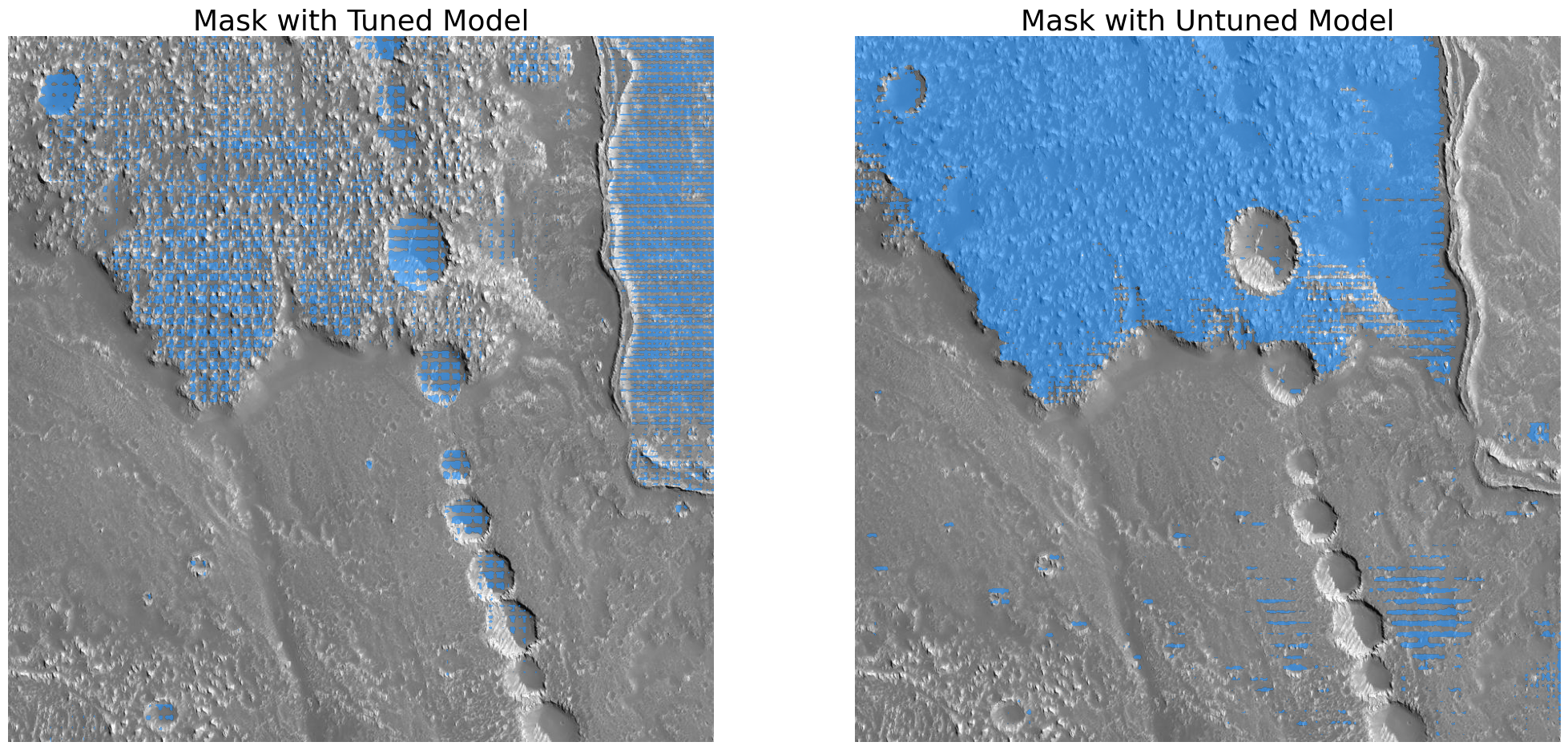}
        \caption{Landforms of interest are harder to detect without prompts while using the SAM decoder. While the untuned model will segment all surrounding regions, the fine-tuned model still struggles with ignoring the regions of non-interest.}
        \label{fig: tuning}
    \end{centering}
\end{figure}

A recent work~\cite{he2023accuracy} from the medical domain corroborates our observation that the model underperforms significantly in comparison to state-of-the-art without training and with just the use of prompts. So fine-tuning the model would be necessary. However, we also observe that the decoder in SAM has learnt patterns other than that specific to the task and is prone to detecting other regions not relevant to our task. In our case, we empirically observed fine-tuned model~\footnote{The SAM decoder is fine-tuned via training with a set of 25 annotated images for 100 epochs.} to give spurious results.~\Cref{fig: tuning} illustrates an exemplary fail-case of fine-tuning the SAM decoder with the labels. SAM decoder even when fine-tuned is optimal only when prompts are available, and thus is hard to be used without human-in-the-loop or additional information from the ground truth. All of the recently developed works~\cite{hu2023sam, ji2023sam, he2023accuracy} use prompts derived from the ground truth labels for the problem-specific task. This is not a realistic scenario in our application. We, therefore, choose to train a separate decoder to learn the problem-specific semantics.

\begin{figure}[h!tb]
    \centering
        \includegraphics[width =0.8\linewidth]{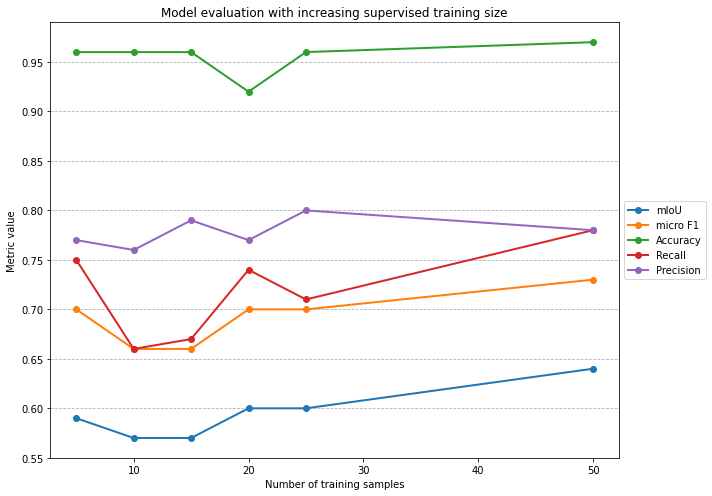}
        \caption{Development of the evaluation metrics with increasing sizes $\increment (\mathbb{N})$ of annotated training samples. Increasing training size beyond a handful of samples yields trivial overall improvement.}
    \label{fig: result_graph}
\end{figure}

\begin{figure}[!htb]
  \centering
  \begin{subfigure}[b]{\textwidth}
    \centering
    \includegraphics[width=0.8\textwidth]{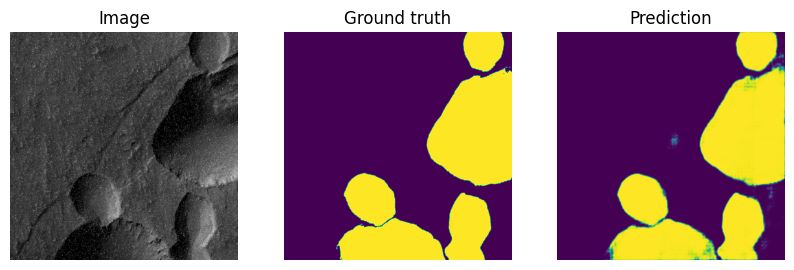}
  \end{subfigure}
  \begin{subfigure}[b]{\textwidth}
    \centering
    \includegraphics[width=0.8\textwidth]{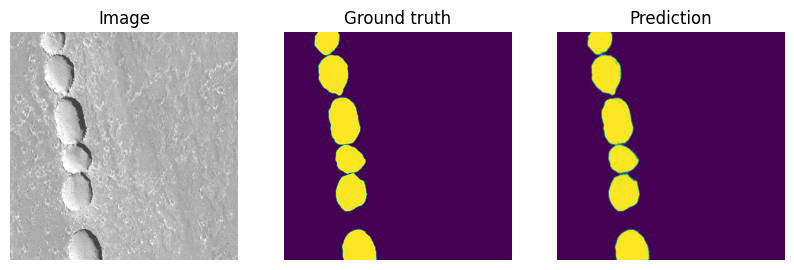}
  \end{subfigure}
  \begin{subfigure}[b]{\textwidth}
    \centering
    \includegraphics[width=0.8\textwidth]{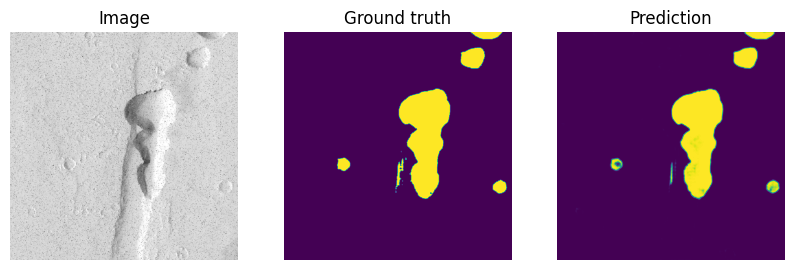}
  \end{subfigure}
    
  \caption{ Example predictions on the test set. The domain decoder identifies all regions of interest reasonably well. }
  \label{fig: example_segmentation}
\end{figure}

We employ a lightweight decoder~(cf.~\Cref{fig: architecture}) comprised of only three upsampling layers using deconvolutions that maps the bottleneck $z$ to an image of the desired size, in our case $(3 \times 900 \times 900)$. The bottleneck is obtained by passing the image through SAMs encoder. During training, only the weights of the decoder are updated. We use a sigmoid activation to map the logits in the range $[0,1]$ for binary segmentation. 
In this manner, we train the decoder with incremental sets of SAM-annotated images. The incremental function $\increment (\mathbb{N})$ is used in step sizes with $\mathbb{N} \in \{5, 10, 15, 20, 25, 50\}$.  All models are trained for a total of 100 training epochs, without additional control.
We compare the performance using mean Intersection over Union (mIoU), micro F1, accuracy, and micro-precision and recall. Micro metrics are chosen to better represent the performance under data imbalance.
~\Cref{fig: result_graph} shows the evolution of the metrics. We observe that the performance improvement with additional training samples after a handful is non-significant, with any differences being representative of stochasticity in evaluation rather than true information gain. By observing the metrics above and with qualitative evaluation it can be inferred that depending on the complexity of the domain-specific task, a very small number of annotations can suffice for a representative performance~(cf.~\Cref{fig: example_segmentation}).

Further, we compare the performance of this approach with $\increment (5)$ against the Mask-RCNN model proposed in existing literature~(cf.~\Cref{tab:arch_comparison}) for the same task, which serves as the benchmark for our comparison. This model is trained with the full training size of 405 manually annotated images. The authors~\cite{nodjoumi2023deeplandforms} in this work only reported macro metrics and noted that about 1000 positive labels were required for satisfactory performance. We clearly see that knowledge distillation through SAM by utilising relatively minuscule labels surpasses the benchmark on most reported metrics. In spite of the precision being slightly lower, the recall is substantially higher. It is to be noted that in tasks like these, recall should be given a higher importance to precision, since missing a region of interest is more critical than falsely identifying one.

\begin{table}[h!tb]
	\vspace*{2mm}
	\centering
	\caption{Comparison of the state of the art vs our proposed approach trained only with 5 labelled samples. The authors in \cite{nodjoumi2023deeplandforms} train their model with 405 samples and report macro metrics.}
	\begin{tabular}{|l||r|r|r|r|}\hline
		\textbf{model} & \textbf{macro F1} & \textbf{accuracy} & \textbf{macro precision} & \textbf{macro recall}  \\\hhline{|=#=|=|=|=|} 
		Mask-RCNN~\cite{nodjoumi2023deeplandforms} & 0.811 & 0.774 & \B 0.952 & 0.706 \\
		\hline
		ours ($\increment (5)$) & \B 0.86 & \B 0.96 & 0.89 & \B 0.93 \\
		\hline
	\end{tabular}
	\label{tab:arch_comparison}
\end{table}

\section{Conclusion}
\label{sec: Conclusion}
In this work, we extended the SAM framework and applied it to the segmentation of landforms like pits and skylights on the surface of Mars using HiRISE images. We observed that SAM has a high accuracy in separating various semantic regions, however, it cannot be directly applied to domain-specific tasks due to a lack of problem-specific bias. To this end, we developed and applied a domain-specific decoder that takes the image embedding generated by SAMs image encoder and learns the problem-specific semantics with substantially fewer labels. By training the domain decoder with only 5 labelled images sampled randomly, we demonstrated an equivalent if not superior performance to the existing Mask-RCNN method for the same task that was trained with over 400 labelled images.

We also explored the applicability of SAMs decoder for annotation using the various out-of-box prompts. We observed that the fully automatic mode is prone to marking irrelevant regions, and further can also miss some regions of interest if it doesn't know where to look. The point-based mode can be ambiguous at times.  In contrast, the bounding box-based mode can clearly specify the ROI and obtain reasonable segmentation results without multiple trials and errors. We can therefore conclude that the bounding box-based segmentation mode can be a useful setting for rapid annotation by the domain expert.

In conclusion, our study reveals that SAM can effectively be exploited to accelerate domain-specific segmentation tasks. This work presents the first attempt to adapt SAM to geological mapping by fine-tuning through knowledge distillation. As part of future work, it might be worthwhile to investigate how the process of annotation can be automated, further lowering the load of human-in-the-loop. We hope this work will motivate more studies to build segmentation foundation models in the planetary science domain.

\bibliographystyle{splncs04}
\bibliography{main}

\end{document}